\documentclass{isprs}
\usepackage{subfigure}
\usepackage{setspace}
\usepackage{geometry} % added 27-02-2014 Markus Englich
\usepackage{enumerate}
\usepackage{amsmath}
\usepackage{amssymb}
\usepackage{natbib}
\usepackage{multirow}  % 多行合并的宏包
\usepackage{stfloats}

\usepackage{graphicx} %Loading the package
\graphicspath{{figs/}}

\usepackage{algorithm}
\usepackage{algorithmic}

\usepackage[x11names]{xcolor}
\usepackage[
     colorlinks,
]{hyperref}

\usepackage{hyperref}
\makeatletter
\newcommand{\printfnsymbol}[1]{%
  \textsuperscript{\@fnsymbol{#1}}%
}
\makeatother

\begin{document}

\title{Geometry-Aware Segmentation of Remote Sensing Images via Joint Height Estimation}

\author{
Xiang Li\textsuperscript{a,b,c}, Lingjing Wang\textsuperscript{a,b,c}, Yi Fang\textsuperscript{a,b,c}\thanks{Corresponding author. Email: yfang@nyu.edu.}
}

\address
{
    \textsuperscript{a } Multimedia and Visual Computing Lab, New York University, New York, United States.\\
    \textsuperscript{b } Tandon School of Engineering, New York University, New York, United States.\\
    \textsuperscript{c } Department of Electrical and Computer Engineering, NYU Abu Dhabi.\\
}

\abstract
{

Deep learning-based methods, especially deep convolutional neural networks (CNNs), have made significant breakthroughs in the field of remote sensing and greatly advanced the performance of the semantic segmentation of remote sensing images. Recent studies have shown the benefits of using additional elevation data (e.g., DSM or nDSM) for enhancing the performance of the semantic segmentation of aerial images. However, previous methods mostly adopt 3D elevation information as additional inputs. While in many real-world applications, one does not have the corresponding DSM information at hand and the spatial resolution of acquired DSM images usually do not match the aerial images. To alleviate this data constraint and also take advantage of 3D elevation information, in this paper, we introduce a geometry-aware segmentation model that achieves accurate semantic labeling of aerial images via joint height estimation. Instead of using a single-stream encoder-decoder network for semantic labeling, we design a separate decoder branch to predict the height map and use the DSM images as side supervision to train this newly designed decoder branch. In this way, our model does not require DSM as model input and still benefits from the helpful 3D geometric information during training. With the newly designed decoder branch, our model can distill the 3D geometric features from 2D appearance features under the supervision of ground truth DSM images. Moreover, we develop a new geometry-aware convolution module that fuses the 3D geometric features from the height decoder branch and the 2D contextual features from the semantic segmentation branch. The fused feature embeddings can produce geometry-aware segmentation maps with enhanced performance. Our model is trained with DSM images as side supervision, while in the inference stage, it does not require DSM data and directly predicts the semantic labels in an end-to-end fashion. Experiments on ISPRS Vaihingen and Potsdam datasets demonstrate the effectiveness of the proposed method for the semantic segmentation of aerial images. The proposed model achieves remarkable performance on both datasets without using any hand-crafted features or post-processing.
}


\maketitle

\section{Introduction}\label{Introduction}
The semantic segmentation problem, which is often called image classification in the field of remote sensing, is generally defined as determining the semantic classes of all pixels in the input images. Automatic semantic segmentation has been a long-standing problem in the field of remote sensing and plays a crucial role in various applications, such as land use/land cover mapping, agricultural production estimate, building extraction, city planning, and etc.

In recent years, convolutional neural networks (CNNs) have drawn huge attention in remote sensing and photogrammetry due to the remarkable performance in many applications, such as scene classification \citep{zou2015deep,cheng2018deep}, image classification \citep{maggiori2016convolutional,marmanis2018classification,audebert2018beyond}, object detection \citep{chen2014vehicle,hu2019sample}, building extraction \citep{mnih2013machine,saito2016multiple,alshehhi2017simultaneous,li2018building}, land use classification \citep{luus2015multiview,castelluccio2015land}, point cloud classification \citep{yang2017convolutional,zhao2018classifying,wen2019directionally}. The encouraging performance drives researchers to develop CNN-based methods for the semantic labeling of remote sensing images (RSIs). In this direction, early efforts adopt patch-based CNNs to predict the class label for the center pixel of each input patch; recent methods mostly perform pixel-wise segmentation using fully convolutional networks. For example, \citep{maggiori2016convolutional} develops a fully convolutional model for the classification of remote sensing images in an end-to-end fashion. Substantial researches have tried to enhance the performance by using more powerfully encoder network \citep{sherrah2016fully,badrinarayanan2017segnet}, incorporating dilated convolution module \citep{zhou2018d,wei2018semantic} or using more powerful output representations \citep{yuan2017learning,chai2020aerial}.

It is commonly known that objects in remote sensing images are characterized by complex spectral-spatial properties and need a comprehensive feature extraction process to ensure the classification performance. Nevertheless, existing CNN-based methods mostly focus on spectral and contextual feature extraction using a single encoder-decoder network, while geometric features (such as height above ground, implicit 3D structure) are often not fully explored. A direct remedy to this issue is to explicitly incorporate geometric-related data (such as DSM) as additional inputs. \cite{audebert2018beyond} propose to enhance the segmentation performance of remote sensing images by fusing the feature representations from both RGB images and elevation composite images (NVDI, DSM, nDSM). Concretely, they propose a two-stream network that simultaneously learns RGB and auxiliary geometric features, and a residual correction module is leveraged to fuse the features from two encoder networks. 

\begin{figure*}[!h]
\centering
\includegraphics[width=12cm]{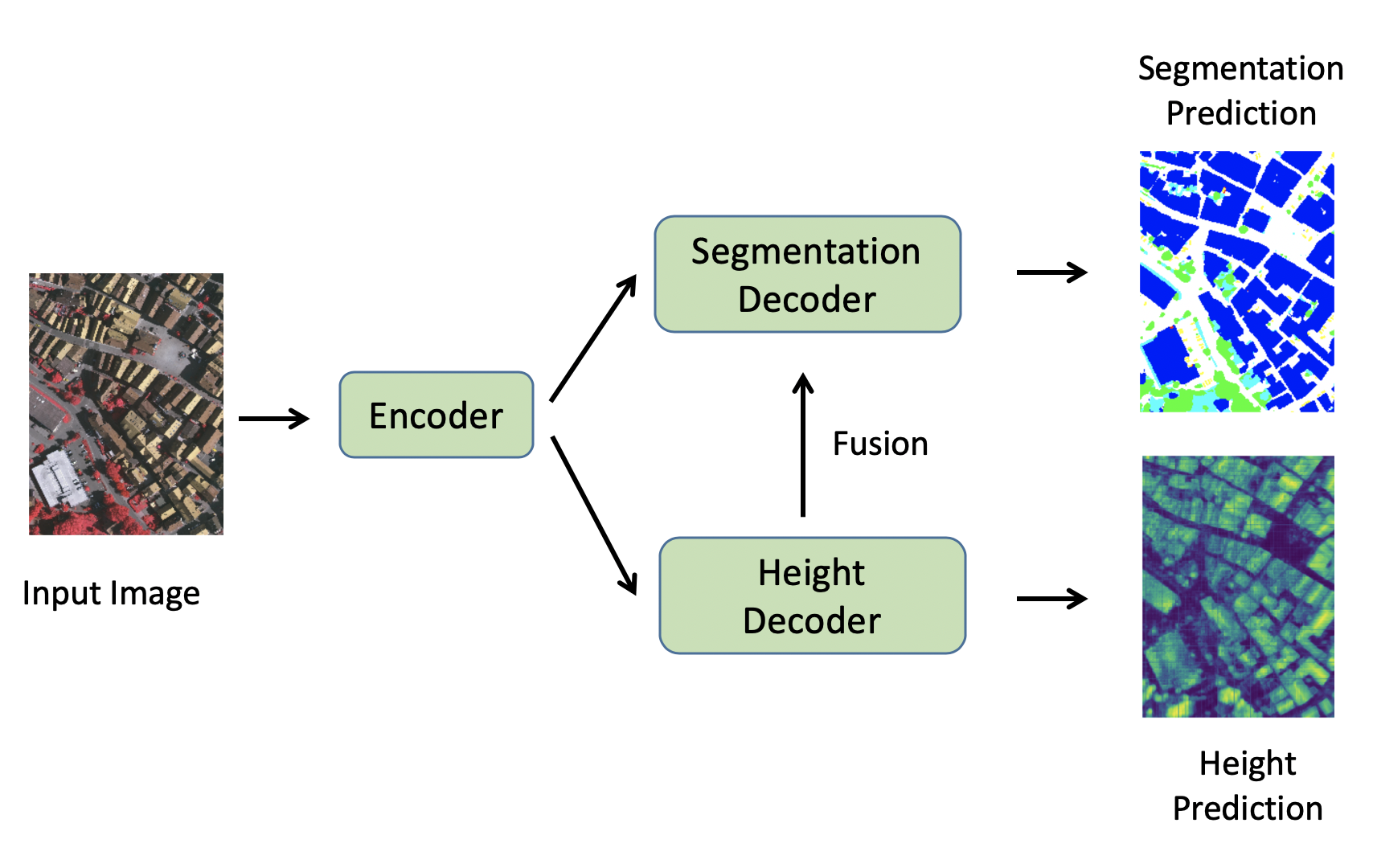}
\caption{
Illustration of our method for simultaneous semantic segmentation and height estimation.}
\label{fig_intro}
\end{figure*}

In this paper, instead of directly taking elevation data (e.g., DSM or nDSM) as additional inputs, we propose to jointly learn geometric features using a height estimation network. Our key insight is that geometric information (height above ground) is naturally preserved by the aerial images and can be estimated from monocular inputs \citep{eigen2014depth, fu2018deep,mou2018im2height,ghamisi2018img2dsm}. The learned 3D geometric features are further fused with the 2D contextual features using the newly designed geometry-aware convolution module. Our model is thus able to distinguish those objects that have similar 2D appearances but with distinct geometric characteristics, e.g., rooftop and impervious surface. Figure \ref{fig_intro} illustrates the proposed framework for simultaneous semantic segmentation and height estimation. Through joint training of these two tasks, the implicit 3D geometric information can be well extracted and fused with contextual features, which further contributes to better semantic labeling performance. More importantly, after training, our model does not need DSM data and can directly produce the segmentation labels for the test images.

We also note that some recent works \citep{Srivastava2017,carvalho2019multitask} explore a multi-task learning strategy for simultaneously height estimation and semantic labeling, which are quite similar to the proposed method. Unlike these methods that decouple two tasks in the middle or top layers of decoder networks, our method uses two task-specific decoder branches, one for semantic labeling and the other for height estimation. More importantly, a geometry-aware convolution module is proposed to effectively fuse semantic feature embeddings and geometric feature embeddings to enhance the performance.

The main contributions of this paper are summarized as follows:

\begin{enumerate}
\item This paper introduces a geometry-aware neural network model for the semantic labeling of aerial images. Instead of taking the DSM images as additional inputs, our model simultaneously predicts the segmentation maps as well as the height maps from input aerial images. After training, it does not need DSM data and can directly produce the segmentation labels for the test images.
\item A geometry-aware convolution module is proposed to effectively fuse semantic feature embeddings and geometric feature embeddings to enhance the performance of semantic labeling.
\item We validate the effectiveness of the proposed method on the ISPRS 2D Semantic Labeling datasets and report remarkable performance compared to the state-of-the-art methods.
\end{enumerate}

The remainder of this paper is organized as follows. In Section \ref{Related Work}, we briefly review the deep learning-based methods for semantic labeling and height estimation on remote sensing images in section \ref{Related Work}. The proposed geometry-aware segmentation model is introduced in Section \ref{Methods}. In Section \ref{Experiments}, we conduct experiments to validate the effectiveness of our method for semantic labeling of remote sensing images. We investigate the effect of our proposed height estimation module and geometry-aware convolution module in Section \ref{sc_discussion}. Finally, our paper is summarized in Section \ref{sc_conclusion}.

\section{Related Work}\label{Related Work}

\subsection{Semantic Segmentation of Remote Sensing Images}\label{sc_ss}
% Semantic segmentation has been a long-standing problem in remote sensing with a lot of both traditional machine learning-based methods and recent deep learning-based methods. In this section, we mainly focused on deep learning-based methods.

% Benefiting from the advances of convolutional neural networks (CNN),  semantic segmentation has obtained significant breakthroughs. Fully Convolutional Networks (FCN)~\citep{long2015fully} stands out as probably the first approach to perform dense predictions by converting fully connected layers into convolutional layers. It thus allows the FCN model to generate segmentation maps for input images of arbitrary sizes. Later, FCN-like methods have achieved great progress in the task of semantic segmentation. In order to recover the object details and spatial information reduced by pooling layers, two typical strategies are exploited. SegNet~\citep{badrinarayanan2017segnet}, UNet~\citep{ronneberger2015u} and RefineNet~\citep{lin2017refinenet} leverage encoder-decoder structure to recover detailed spatial information by fusing low-level and high-level layers step-by-step. On the other hand, dilated convolution is used in ~\citep{yu2015multi,chen2014semantic,chen2018encoder,zhao2017pyramid} to reserve high-resolution feature maps while enlarging the receptive field of the neural network.

Benefiting from the powerful feature learning abilities of deep neural networks, semantic segmentation of remote sensing images have achieved significant improvements. In this direction, \citep{mnih2010learning} is the first successful work that utilizes a patch-based CNN model for road and building extraction. \cite{saito2016multiple} develop a method for simultaneous road and building extraction on an aerial imagery dataset using a single CNN network. A moving-average technique is designed to ensemble model predictions with different spatial displacements and further enhance the performance. \cite{vakalopoulou2015building} further extend this method to multi-spectral images and validate the performance of their model for building extraction. Nevertheless, these patch-based methods need to divide the original input images into small patches and can only produce one classification label for each patch. They need to slide over the whole image plane to get the final prediction, which makes these methods inefficient for large-scale datasets. 

A breakthrough comes from Fully Convolutional Network (FCN) \citep{long2015fully} that nicely converts the fully connected layers of CNNs into convolutional layers, and therefore enables dense pixel-wise segmentation of input images. After FCN, numerous variants have been proposed to enhance the performance by using more powerful encoder networks, skip-connections, or dilated convolution modules. In remote sensing, recently proposed classification methods are mostly based on an encoder-decoder architecture. For example, \citep{maggiori2016convolutional} develop a fully convolutional architecture for the pixel-wise classification of remote sensing images. \cite{marmanis2016semantic} design an FCN to perform pixel-wise classification on the ISPRS semantic labeling benchmark. They also discuss different design choices of the proposed method and demonstrate an ensemble of CNNs can achieve better results. %\citep{sherrah2016fully} proposes to infer a full-resolution labeling map using a deep FCN with no downsampling layers, avoiding the need for deconvolution or upsampling layers. 
\citep{marmanis2018classification} explicitly adds a boundary detection branch to the SegNet~\citep{badrinarayanan2017segnet} architecture to preserve high-frequency details of object boundaries. They show that adding a boundary detection network can enhance the semantic labeling performance of remote sensing images.
Instead of focusing on the design of network architecture, other researchers try to boost performance by using more powerful output representations. \cite{yuan2017learning} proposes a novel building extraction method that uses the signed distance from each pixel to building boundaries to represent output and report significant performance boost over the baseline model using traditional label maps. A more recent work \citep{chai2020aerial} extends this method to the multi-class classification of aerial images and train a CNN model to predict multiple signed distance maps, one for each class. 

Moreover, some recent work also tries to improve performance by taking advantage of multi-modal fusion. \cite{liu2017dense} develop two-stream network for VHR image segmentation. An FCN branch is used to generate classification probabilities from optical images; meanwhile, a multinomial logistic regression branch is used to generate probability maps from LiDAR inputs. A high-order CRF model is leveraged to fuse these two probabilistic results. \cite{audebert2018beyond} propose a two-stream network that jointly learns RGB and depth features, and they also investigate the early and late fusion of two sources of input images. Instead of using DSMs as additional inputs, \citep{volpi2018deep} proposes to treat DSMs as mid-level features and feed them to the hypercolumn layer of a VGG network.

\subsection{Height Estimation from Single Aerial Images}\label{sc_related_height}

In the field of remote sensing, existing methods for height estimation mostly focus on 3D reconstruction based on stereo or multi-view image matching. There are only a few numbers of researches focus on estimating height from single aerial images. Early efforts mostly start by identifying object shades based on pixel-wise or object-based features and then estimate height values using camera information \citep{shao2011shadow,comber2012using,kim2007semiautomatic}. For example, \citep{kim2007semiautomatic} propose a building height estimation method that firstly projects the building shadow onto the ground plane and then adjusts the building height until the projected shadow can be well-aligned with the real one. Much attention has been paid to enhance the shadow detection performance by using more powerful contextual and geometric features \citep{shao2011shadow,comber2012using}. Other researches also try to estimate height from single remote sensing images using a small number of control points \citep{chen2012densification} or DTMs \citep{rajabi2004optimization}.

Height estimation from aerial images is similar to the task of depth estimation in the field of computer vision. Recent progress has frequently shown the capabilities of deep neural networks for learning representative depth cues from single RGB images \citep{eigen2014depth,eigen2015predicting,kuznietsov2017semi,laina2016deeper,li2017two}. In \citep{eigen2014depth}, the authors introduce a multi-scale CNN network to predict depth from monocular RGB images. The proposed method includes two branches for coarse-to-fine prediction: the global branch uses fully connected layers to get the coarse depth map, while the local branch use fully convolutional networks to refine the coarse depth predictions. The following works, such as \citep{laina2016deeper,fu2018deep}, try to improve the performance by using more powerful network architecture. Other methods combine CNNs with probabilistic graphic models (e.g., CRF or MRF) to refine the pixel-level depth estimations \citep{liu2015learning,wang2015towards}. More recent researches also try unsupervised or semi-supervised methods for monocular depth estimation \citep{garg2016unsupervised,kuznietsov2017semi}. 

In the light of CNN-based methods for monocular depth estimation, some recent researches have explored CNN-based models for height estimation with single aerial images as inputs. In \citep{mou2018im2height}, the authors train an encoder-decoder network to predict the height map from a single aerial image. \cite{amirkolaee2019height} also adopts an encoder-decoder network for height estimation and further introduces a post-processing technique to generate absolute DSMs from overlapped relative height predictions. \cite{ghamisi2018img2dsm} adopts a generative adversarial network (GAN) to predict height maps from high-resolution aerial images and naturally enforces the generated height maps to have the same distribution with real ones.

\subsection{Multi-task Learning}
The proposed model can be considered as one of the methods based on multi-task learning. Multi-task learning aims at solving multiple tasks simultaneously by exploiting commonness and differences among these related tasks. Some recent works try multi-task learning for simultaneously semantic labeling and height estimation from remote sensing images. In \citep{Srivastava2017}, the authors develop a multi-task CNN model for simultaneous height estimation and semantic segmentation. The proposed method uses a shared encoder-decoder backbone network for high-level task-agnostic feature learning, and two task-specific heads to predict classification map and height map simultaneously. The most similar work to our method comes from \citep{carvalho2019multitask}. In \citep{carvalho2019multitask}, a multi-task learning-based architecture is also proposed for simultaneous height estimation and classification of remote sensing images. In this method, the authors explore the middle-level split in the decoder network instead of doing it in the very top layers. In contrast to \citep{carvalho2019multitask}, our proposed method decouples the two tasks right after the encoder network, and design two decoder branches to enable more representative task-specific feature learning. More importantly, our model introduces a geometry-aware convolution module to further fuse the high-level semantic and geometric features in two decoder branches to enable geometry-aware semantic labeling.

\section{Methods}\label{Methods}
In this section, we introduce our geometry-aware semantic segmentation model. First, we give an overview of the proposed method in Section \ref{sc_overview}. The encoder-decoder network is introduced in Section \ref{sc_feature}. The proposed geometry-aware convolution module is illustrated in Section \ref{sc_gac}. The multi-task loss function is presented in Section \ref{sc_loss}.

\begin{figure*}[t]
\centering
\includegraphics[width=17cm]{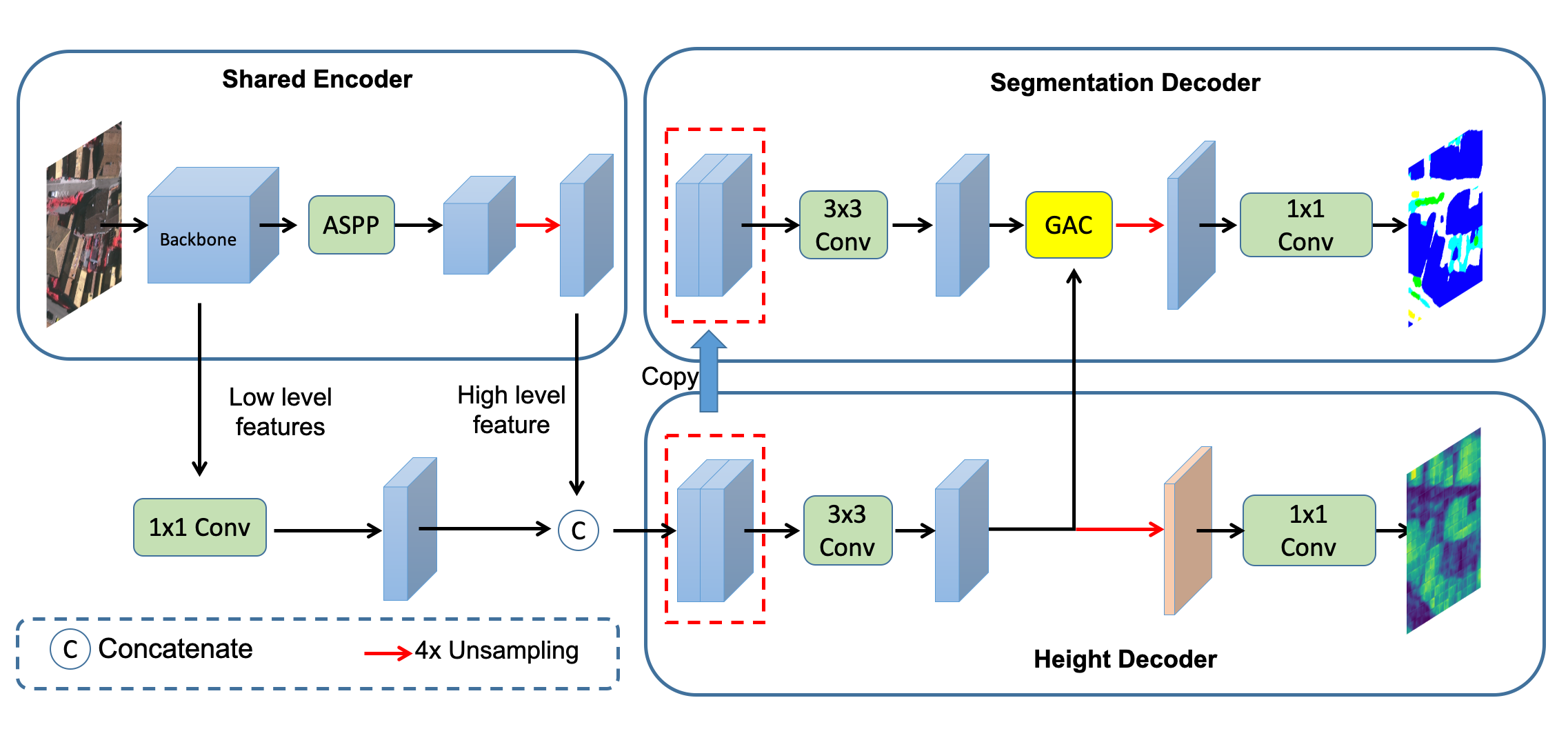}
\caption{
Overview of our GANet model for remote sensing image semantic labeling. Our model receives a single aerial image as input and predicts the classification map and height map simultaneously. The shared encoder adopts ResNet-101 as a backbone network to extract high-level features from input images. An atrous spatial pyramid pooling (ASPP) module is appended after the backbone network to fuse multi-scale features. The combined feature maps are upscaled by 4x and further concatenated with the low-level features of the same resolution from the backbone network.
In the decoder part, the combined feature maps are fed into two separate $3 \times 3$ convolution layers to learn independent feature representations for semantic segmentation and height estimation, respectively.
A geometry-aware convolution module is then used to further fuse these two form of features thus enables geometry-aware semantic labeling.}
\label{fig_overview}
\end{figure*}

\subsection{Method Overview}\label{sc_overview}
Given a group of aerial images $\mathcal{I}=\{\mathcal{I}_1, \mathcal{I}_2, ...,\mathcal{I}_{N}\}$ and corresponding label maps $\mathcal{Y} = \{\mathcal{Y}_1, \mathcal{Y}_2, ..., \mathcal{Y}_{N}\}$, where $\mathcal{I}_i \in \mathcal{R}^{H \times W \times3}$ and $\mathcal{Y}_i \in \mathcal{R}^{H \times W}$. $H$ and $W$ indicate the image height and width respectively. Our semantic segmentation model aims to predict a classification map $\hat{\mathcal{Y}}_i$ for each input image $\mathcal{I}_i$. Traditional single-stream encoder-decoder based networks use a successive of convolutional and pooling layers to obtain high-level contextual features from input images, and then a successive of convolutional and unpooling layers are adopted to decode the learned features into classification score maps. A pre-defined classification loss, e.g., cross-entropy loss, formulated on the predicted classification maps and the ground truth ones are used to optimize the network parameters. In this way, the network can learn only 2D contextual/appearance features, while neglecting the 3D geometric information which is also important for distinguishing those objects that have similar 2D appearances but with different geometric characteristics, e.g., rooftop and impervious surface. 

In this paper, our proposed method explicitly enables geometric feature learning by incorporating a new decoder branch. During training, the 3D information from ground truth height maps is used to guide the training procedure of the newly designed decoder branch. %In this way, our model distill out the geometric feature learning from the contextual feature learning.
Figure \ref{fig_overview} illustrates the proposed Geometry-Aware segmentation network (GANet) for aerial image classification. Our GANet model contains three main components: the encoder network, segmentation decoder, and height decoder. The encoder network aims to learn both contextual and geometric features from input images, which will be introduced in Section \ref{sc_feature}. The segmentation decoder predicts classification maps while the height decoder learns the geometric embeddings by predicting height maps. 
% Features from the height decoder are propagated (by summation) to the segmentation decoder to provide multi-scale geometric guidance. 
After getting the contextual and geometric feature embeddings, a geometry-aware convolution module (GAC) is used to fuse these two forms of features to enables geometry-aware semantic labeling. The GAC module is illustrated in Section \ref{sc_gac}.

\subsection{Encoder-Decoder Network}\label{sc_feature}
\begin{figure}[t]
\centering
\includegraphics[width=8cm]{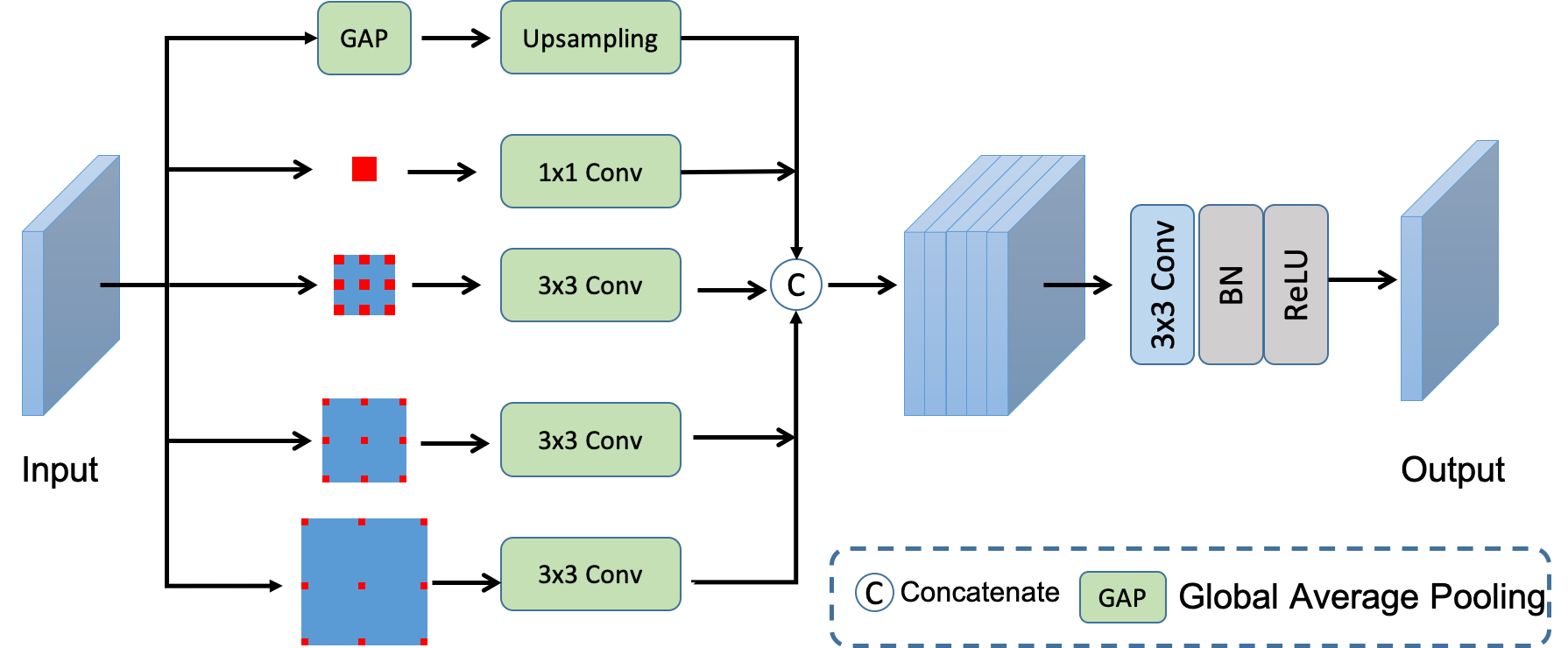}
\caption{
Illustration of our ASPP module.}
\label{fig_aspp}
\end{figure}

Our GANet follows the prevalent Deeplab V3+ \citep{chen2018encoder} architecture to design its encoder and decoder parts. In the encoder part, a backbone network (e.g., VGG-16, ResNet-101) is used to extract multi-scale feature representations. An atrous spatial pyramid pooling (ASPP) module is applied after the backbone network to learn multi-scale features. In our method, the ASPP module consists of one regular convolutional layer and three dilated convolutional layers with a dilated rate of 6, 12 and 18 respectively. A global average pooling layer is also leveraged in the ASPP module to encode full-image information and is further up-sampled to the original resolution. Figure \ref{fig_aspp} gives an illustration of the ASPP module used in this paper. The multi-scale feature representations after the ASPP module are then upscaled by 4x and further concatenated with low-level features of the same resolution. Note that the low-level features are fed into another convolution layer before concatenation. \\

In the decoder part, the combined features are fed into two separate $3 \times 3$ convolution layers to learn independent feature representations for the task of semantic segmentation and height estimation, respectively. The learned height-related geometric feature embeddings are directly upscaled by 4x and fed into another convolutional layer to predict the height maps. The learned semantic-related contextual features are fused with the geometric features by leveraging the newly proposed geometry-aware convolution (GAC) module for enhancement. The GAC module is introduced in the next section. The fused feature maps are then passed to a convolutional layer to predict the semantic labels. In this paper, our feature extraction network is build upon ResNet-101 \citep{he2016deep} architecture and is pre-trained on PASCAL VOC 2012 dataset \citep{everingham2015pascal}.

\subsection{Geometry-Aware Convolution Module}\label{sc_gac}

% After getting contextual and geometric feature embeddings, we fuse these two features through a newly designed convolution module. 
Before introducing the newly proposed convolution module, let's revisit conventional convolutional layer first. A conventional 2D convolution operation aggregate all activations within a local neighborhood. Given an input feature map $X \in \mathbb{R}^{H \times W \times C}$, where $H$ and $W$ denote the width and height of the feature map, $C$ denotes the number of input channel, the output feature vector at pixel $i$ can be calculated as:
\begin{equation}
y_i = \sigma(\sum_{j \in \mathcal{N}_i} W_{ij} x_j + b)
\end{equation}
where $x_j \in \mathbb{R}^C$ denotes the input feature vector at neighbor pixel $j$, $\mathcal{N}_i$ denotes the local neighborhood of pixel $i$, and $y_i \in \mathbb{R}^E$ ($E$ is the output dimension) denote the output feature vector at pixel $i$, $\sigma$ denotes the activation function (e.g., sigmoid), $W \in \mathbb{R}^{C\times E}$ is the convolution kernel which is shared across all pixel locations, $b \in \mathbb{R}^E$ denotes the bias.

Previous researches \citep{wang2018depth,chen20193d} have explored the geometric correlations between pixels by adding a geometric similarity term to the convolution operation. In \citep{wang2018depth}, the author propose a depth-aware convolution operation calculated as:
\begin{equation}
y_i = \sigma(\sum_{j \in \mathcal{N}_i} S(d_i, d_j) W_{ij} x_j + b)
\end{equation}
where $d_i$ and $d_j$ denote the depth values at pixel location $i$ and $j$ respectively, $S(\cdot, \cdot)$ measures the depth similarity between two depth values. By this formulation, the neighbor pixel which has a similar depth with the center pixel $i$ will have a larger impact on the convolution output.

Motivated by this formulation, we introduce a geometry-aware convolution module that leverages the learned geometric embeddings as guidance for the convolution operation. Instead of using original height values as convolution inputs, the proposed convolution operation takes as input both contextual and geometric features in the embedding space. Given an input contextual feature map $x$ and the learned geometric embeddings $G \in \mathcal{R}^{H \times W \times E}$, the convolution output $y_i$ at location $i$ can be formulated as:
\begin{equation}
y_i = \sigma(\sum_{j \in \mathcal{N}_i} W_{ij}(G) x_j + b)
\end{equation}
where $W_{ij}$ is the kernel weights derived from the geometry guidance $G$. Here $W_{ij}$ can be regarded as a geometric similarity between pixel $i$ and $j$ defined in the embedding space. To better calculate $W_{ij}$, we follow \citep{jiao2019geometry} to decouple it as a dot-product of two sub-space embeddings:
\begin{equation}
W_{ij}(G) = \phi(G_i) \cdot \psi(G_j)
\end{equation}
where $\phi(\cdot)$ and $\psi(\cdot)$ denote the features in sub-embedding space. Then, the proposed geometry-aware convolution operation is defined as:
\begin{equation}
y_i = \sigma(\sum_{j \in \mathcal{N}_i} \phi(G_i) \cdot \psi(G_j) \cdot x_j + b)
\end{equation}
Note that our proposed convolution operator is close to the Non-local Neural Networks presented in \citep{wang2018non}. In \citep{wang2018non}, the convolution operator takes as input a single feature map; while in our convolution module, the inputs combines the feature embeddings from both semantic space and geometric space.

\begin{figure}[t]
\centering
\includegraphics[width=8cm]{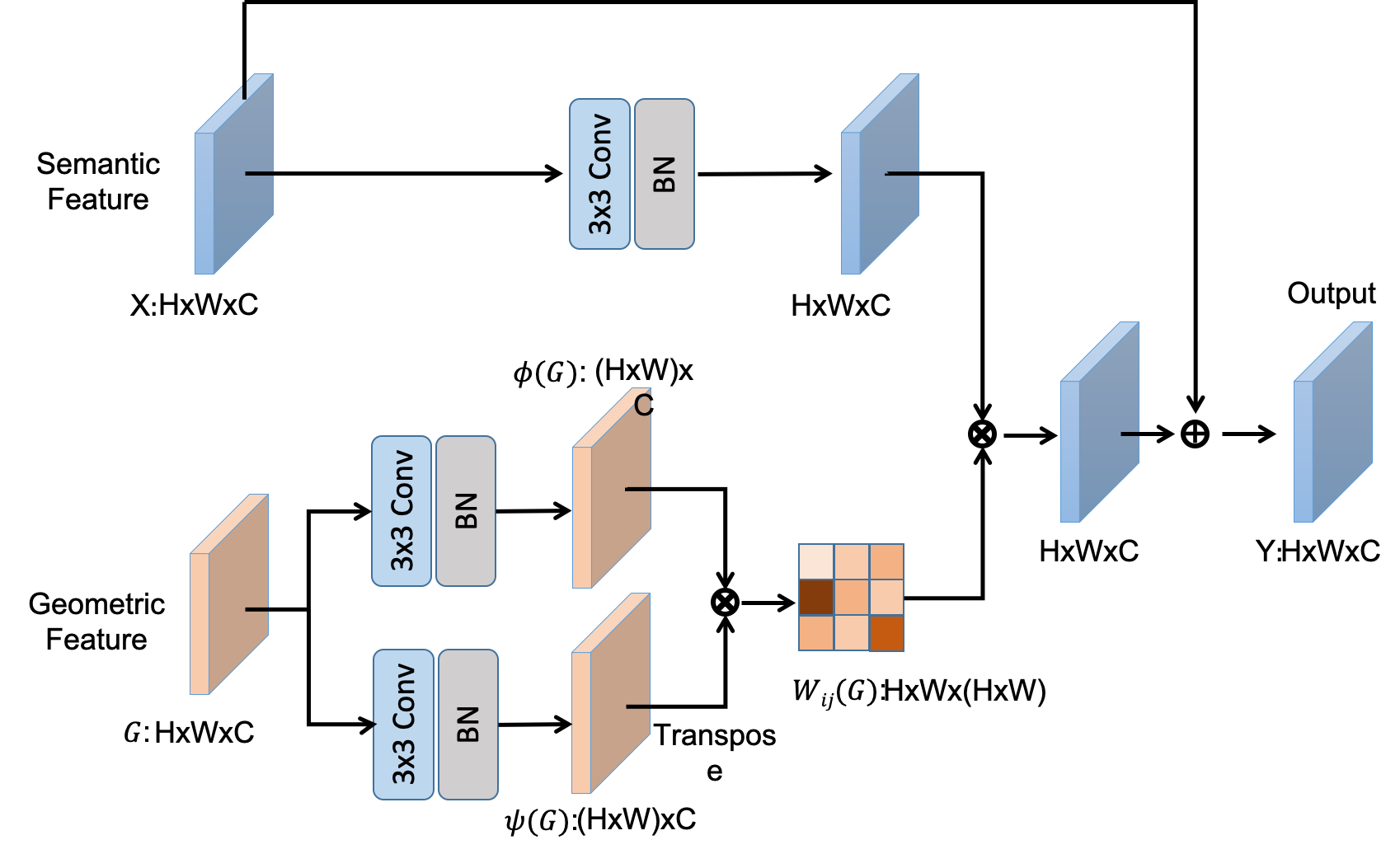}
\caption{
Illustration of geometry-aware distillation module. $\bigotimes$ represents
dot-product and $\bigoplus$ represents element-wise summation.}
\label{fig_gac}
\end{figure}

In this paper, we achieve this convolution operator by several conventional convolutional layers and several basic element operations. Figure \ref{fig_gac} gives an illustration of the proposed GAC convolution module. The geometric embeddings are first passed to two convolutional layers to get the sub-embeddings $\phi(G)$ and $\psi(G)$. Then the geometric similarity matrix $W_{ij}(G)$ is produced by dot produce of two sub-embeddings feature maps. After that, the geometric affinity is fused with semantic features by another dot product operation. Finally, the fused information is combined with original contextual features through an element-wise summation to get the final outputs. Note that the whole convolution process maintains the dimension and size of the contextual features.

\subsection{Multi-task Objective Function}\label{sc_loss}
Our GANet model gets supervision from both semantic segmentation branch and height estimation branch. The overall loss function is formulated as:
\begin{equation}
    \mathcal{L} = \mathcal{L}_{seg} + \lambda \mathcal{L}_{g}
\end{equation}
where $\mathcal{L}_{seg}$ denotes the segmentation loss and $\mathcal{L}_{h}$ denotes height estimation loss, $\lambda$ is a hyper-parameter to balance these two loss terms. By default, $\lambda$ is set to 1 in our experiments.

For the semantic segmentation task, existing methods mostly use cross-entropy loss to penalize the difference between the ground truth labels and predicted label maps. In this paper, we note that in remote sensing datasets, different semantic classes can have a very different number of pixels (e.g., the car category has a much smaller number of pixels than the vegetation category). To address this issue, we leverage the weighted cross-entropy loss for model training, where the inverse class frequencies are used as the balance weights for all pixels of that class. Our semantic segmentation loss function can be calculated as follows:
\begin{equation}
\mathcal{L}_{seg} = \sum_{i} w_i \sum_{c} \ell_{i} log(p_{ic})
\end{equation}
where $i$ indicates the pixel location, $c$ denotes the category index, $\ell_i$ is the ground truth label of pixel $i$, $p_{ic}$ is the predicted probability of pixel $i$ belonging to class $c$, $w_i$ denotes the balance weight for pixel $i$.

% In addition to the pixel-wise classification supervision, learning the geometry-aware embeddings requires supervision from the height domain. 
In this paper, we adopt L1 loss to train height estimation network, calculated as:
\begin{equation}
\mathcal{L}_h = \sum_{i}|\hat{H_i} - H_i|
\end{equation}
where $\hat{H_i}$ and $H_i$ denote the predicted and ground truth height at pixel $i$.

\begin{figure*}[]
\centering
\subfigure[Aerial image.]{\includegraphics[width=4.3cm]{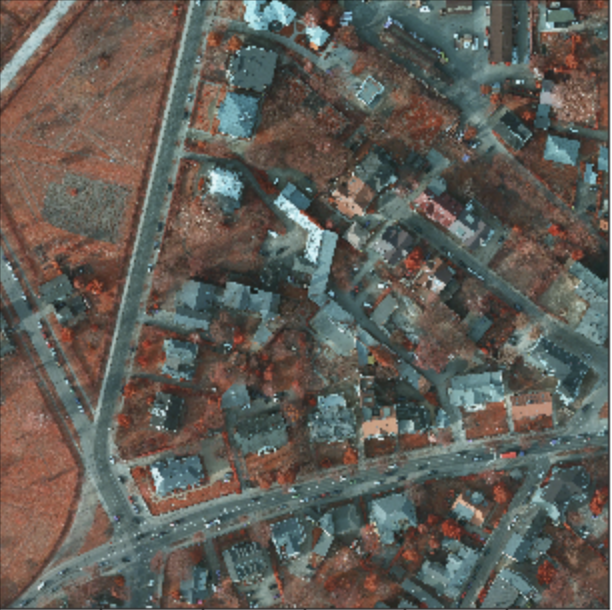}}\hspace{2pt}
\subfigure[DSM.]{\includegraphics[width=4.3cm]{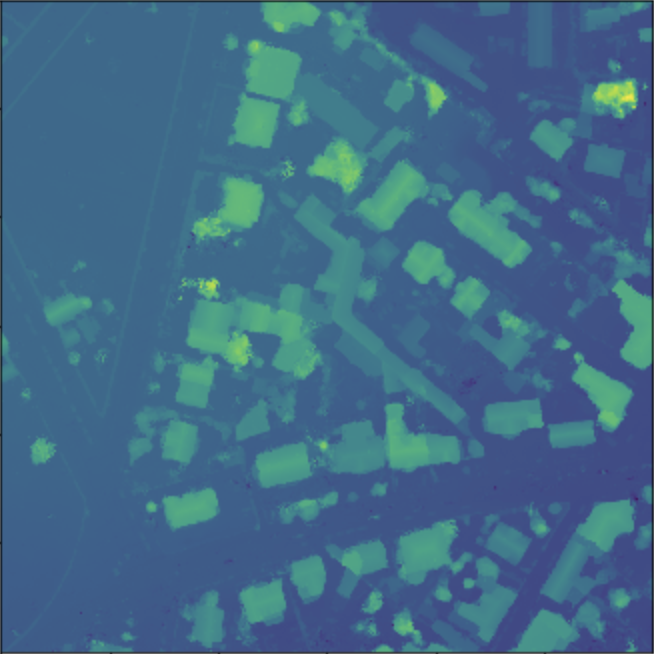}}\hspace{2pt}
\subfigure[Ground Truth.]{\includegraphics[width=4.3cm]{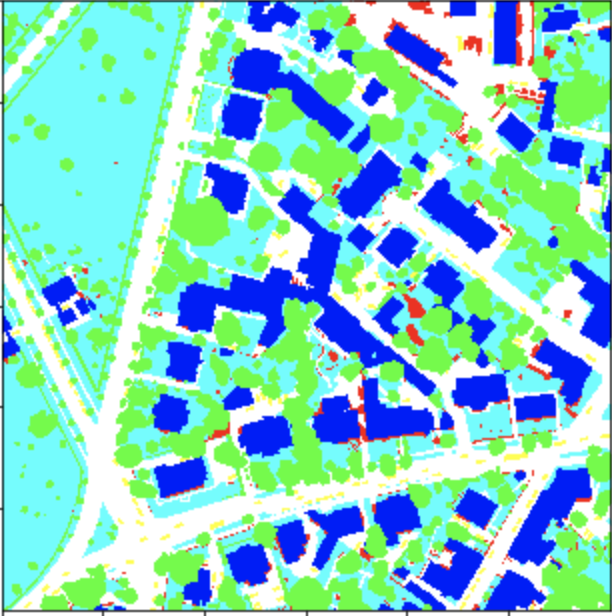}}\hspace{2pt}

\caption{Examples from ISPRS Potsdam dataset. From left to right: (a) aerial image, (b) corresponding DSM image, (c) ground truth label map.}
\label{fig_dataset}
\end{figure*}

\section{Experiments and Results}\label{Experiments}

\subsection{Datasets}\label{sc_dataset}
To verify the effectiveness of our proposed model for the semantic labeling of remote sensing images, we conduct experiments on the ISPRS 2D Semantic Labeling Challenge dataset \citep{niemeyer2014contextual}. This dataset contains very high-resolution aerial images from two cities of Germany: Vaihingen and Potsdam. And for each aerial image, the ground truth labels are provided on six classes: buildings, impervious surfaces (e.g., roads), trees, low vegetation, cars, and clutter. The corresponding DSM information generated by dense image matching is also provided.

\subsubsection{ISPRS Vaihingen}
The Vaihingen dataset includes 33 image tiles at a spatial resolution of 9cm/pixel, and each tile has around $2500\times2500$ pixels. Each aerial image comprises three channels of near infrared, red, and green. Following the official split, 16 tiles with provided ground truth are used for model training, and the remaining 17 tiles are used for held-out evaluation by the challenger organizers. Among the training set, four tiles (image numbers 5, 7, 23, 30), are selected as the validation set.

\subsubsection{ISPRS Potsdam}
The Potsdam dataset includes 38 image tiles at a spatial resolution of 5cm/pixel, and each tile has $6000\times6000$ pixels. Each aerial image comprises four channels of near infrared, red, green, and blue. Following the official split, 24 tiles with provided ground truth are used for model training, and the remaining 14 tiles are used for held-out evaluation by the challenger organizers. Four tiles (image numbers 7\_8, 4\_10, 2\_11, 5\_11) from training split are selected as the validation set to determine the optimal hyper-parameter configurations. Figure \ref{fig_dataset} shows an example of the aerial image, corresponding DSM image, and label map from the ISPRS Potsdam dataset.

\subsection{Implementation details}
Our GANet model is implemented based on PyTorch Library. The network is optimized using a momentum SGD algorithm with the momentum set to 0.9. We train our model for 100 epochs with a cosine learning rate decay schedule. The initial and minimum learning rate is set to 0.01 and 0.00002, respectively. We use a weight decay of 0.0005 for regularization. We train our model on 4 Tesla P100 GPUs with the batch size set to 4. We use Synchronized BN \citep{zhang2018context} after each convolutional layer.

Considering each image tile in both Vaihingen and Potsdam datasets has quite a large size, it can not be directly used for model training due to the GPU memory limit. An intuitive remedy to this issue is to divide the original image tiles into small patches. Previous methods mostly use a sliding window strategy to extract image patches. In this paper, we randomly select a patch of size $320 \times 320$ pixels ($512 \times 512$ pixels for Potsdam dataset) from each tile to formulate the training batch. By doing so, the input patches can be chosen from all possible positions in the image tile instead of the pre-defined locations when using a sliding window strategy. For data augmentation, we randomly flip the training patches horizontally or vertically with a probability of 0.5.

We select the best hyperparameter configurations based on the performance on the validation dataset and use it for online evaluation. In the test stage, a sliding window approach is used to generate small patches of size $320 \times 320$ pixels ($512 \times 512$ for the Potsdam dataset). We set the sliding stride to 32 pixels to ensure overlaps between consecutive patches. We feed all image patches into our trained model and generate the probability map for each patch and average the probability values in overlapping regions. Considering objects in aerial images can have very different sizes, we use multi-scale inputs (scales of 0.8, 1, 1.2) to enhance the testing performance.

\subsection{Evaluation metric}
We use overall accuracy (OA) and per-class F1 score to evaluate the performance of our GANet model. The OA evaluates the classification performance by the percentage of correctly classified pixels over all pixels. The F1 score is defined as the harmonic average of precision and recall of a given class and is calculated as follows,
\begin{equation}
precision=  \frac{TP}{TP+FP}\\
\end{equation}
\begin{equation}
recall =  \frac{TP}{TP+FN}\\
\end{equation}
\begin{equation}
F1 = \frac{2*precision*recall}{precision+recall}\\
\end{equation}
, where $TP$ denotes the number of true positive pixels, $FP$ denotes the number of false positive pixels and $FN$ denotes the number of false negative pixels. 

Moreover, to lower the effect of uncertain border definitions during the evaluation, we follow contest protocols to erode object boundaries by a 3-pixel circle and ignoring those pixels during evaluation.

\subsection{Comparing methods}
In our experiments, we compare our GANet model with methods submitted to the ISPRS 2D Semantic Labeling Contest. The implementation details of each comparing method are listed below.\\
1) SVL\_*: This is the baseline method provided by the challenge organizer \citep{gerke2014use}. This method takes as input both aerial images and several additional pre-calculated features, including  SVL-features \citep{gould2009stair}, normalized digital vegetation index (NDVI), saturation, and normalized DSM (nDSM). The Adaboost algorithm is used to get the initial results, and the conditional random field (CRF) model is used as post-processing. \\
2) UZ\_1: The method is developed by \citep{volpi2016dense}. It takes as input both aerial images and nDSM images. This method uses an encoder-decoder network in which a successive of convolutional layers are used to learn high-level features and then deconvolution layers are used to predict the final classification outputs.\\
3) ADL\_3: The method is developed by \citep{paisitkriangkrai2016semantic}. The model input combines aerial images, nDSM, and nDSM data. In this method, both CNN and Random forest (RF) classifiers are used to produce per-pixel classification probability maps, and the results are fused. CRF is applied to further improve performance.\\
4) DST\_2: The method is developed by \citep{sherrah2016fully}. It takes as input both aerial images and DSM data. A deep FCN model is proposed to obtain per-pixel labels with no downsampling and upsampling layers. CRF is applied to further improve performance.\\
5) DLR\_10: This method is developed by \citep{marmanis2016semantic}. It takes as input both aerial images and DSM data. This method combines classification with edge detection in an end-to-end network using a SegNet-based architecture.\\
6) ONE\_7: This method is developed by \citep{audebert2016semantic}. This method contains two multi-scale branches, one branch is trained with aerial images, and the other is trained with the composite image of NDVI, DSM, and nDSM.\\
7) RIT\_2: This method is developed by \citep{piramanayagam2016classification}. In this method, two independent convolutional layers are adopted to extract representative features from RGB images and composite images (IR, NDVI, and nDSM), respectively. The output feature maps are fused and fed into an FCN network for pixel-wise labeling.\\
8) SWJ\_2: This method is developed by \citep{wang2019deep}. It uses the ResNet-101 backbone network for high-level feature extraction and builds a fully convolutional network that adaptively fuses multi-scale features. Only IRRG images are used as the inputs for model training and evaluation. \\
9) CASIA\_2: This method is developed by \citep{liu2018semantic}. It only takes aerial image data as input. This method is based on ResNet-101 architecture, and it does not require the elevation data (DSM and nDSM) or any post-processing techniques.\\
10) TreeUNet: This method is developed by \citep{yue2019treeunet}. This method directly takes the aerial image data as well as DSM data as inputs for model training. No hand-crafted features or post-processing are used.

\subsection{Results on Vaihingen}

\begin{figure*}[!t]
\centering
\includegraphics[width=17cm]{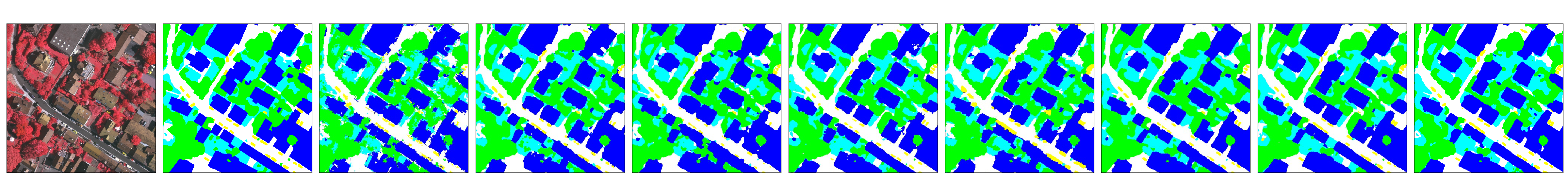}\hspace{2pt}
\includegraphics[width=17cm]{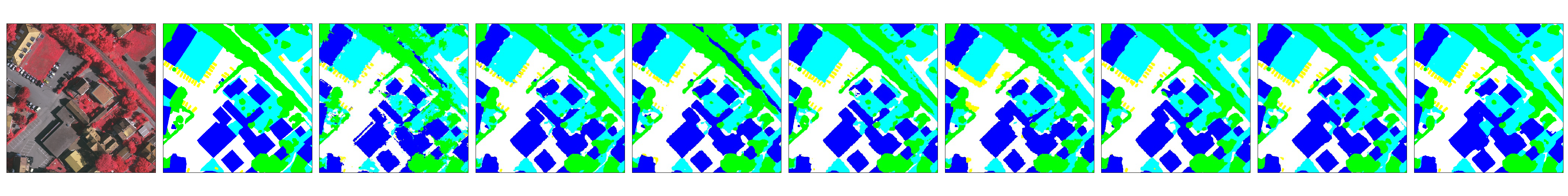}\hspace{2pt}
\includegraphics[width=17cm]{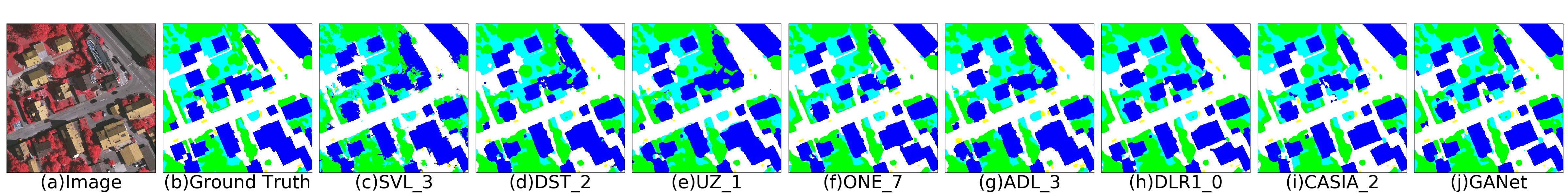}\hspace{2pt}
\caption{Selected examples of the segmentation results on ISPRS Vaihingen dataset.}
\label{fig_rst_vaihingen}
\end{figure*}

The quantitative performance of our GANet model on the test split of ISPRS Vaihingen dataset in Table \ref{tab_result_vaihingen}. In this table, we compare our GANet model with other best-published methods on the ISPRS Vaihingen challenge. As indicated in Table \ref{tab_result_vaihingen}, our GANet model gets better performance than all comparing methods with an OA of 91.3\% and an average F1 score of 90.4\%. The best-published CASIA2 model achieves quite close performance as our model. Note that CASIA2 pre-trains their model on PASCAL VOC 2012 dataset and then finetunes their model on the ISPRS Vaihingen dataset. Ablation analysis shows that CASIA2 model improves the performance a lot by using the finetune technique. Our GANet model is trained from scratch and does not need pretraining from other datasets. %Moreover, our GANet model achieves new state-of-the-art F1 scores on 3 out of 5 categories, including low vegetation, tree, and car.
%Moreover, our proposed GANet model achieves a comparable performance with CASIA2 on average F1 score.

Figure \ref{fig_rst_vaihingen} shows the classification results of our GANet model and the compared methods on several sampled patches. As can be seen in this figure, our GANet model produces satisfying classification results on all test samples. Moreover, our model can better distinguish between the building and impervious surface categories, as well as vegetation-tree categories. We owe this to the 3D geometric difference between these objects.

\begin{table*}[!h]
    \centering
    \caption{Classification performance on the Vaihingen dataset. `DSM(s)' indicates the model using DSM as additional supervision}
    \begin{tabular}{c c c c c c c c c}
        \hline
        Method & Input & Imp. surf. & Buildings &    Low veg. &    Trees &    Cars &    OA & Average F1\\
        \hline
        SVL\_3 & IRRG+nDSM+NDVI+SVL & 86.6 & 91.0 & 77.0 & 85.0 & 55.6 & 84.8 & 79.0 \\
        DST\_2 & IRRG+DSM &90.5 & 93.7 & 83.4 & 89.2 & 72.6 & 89.1 & 85.9\\
        UZ\_1 & IRRG+nDSM & 89.2 & 92.5 & 81.6 & 86.9 & 57.3 & 87.3 & 81.5\\
%         RIT_L7 & 90.1 & 93.2 & 81.4 & 87.2 & 72.0 & 87.8 & 84.8
        ONE\_7 & IRRG+DSM+NDSM & 91.0 & 94.5 & 84.4 & 89.9 & 77.8 & 89.8 & 87.5\\
        ADL\_3 & DSM+nDSM & 89.5 & 93.2 & 82.3 & 88.2 & 63.3 & 88.0 & 83.3\\
        DLR\_10 & IRRG+DSM+Edge & 92.3 & 95.2 & 84.1 & 90.0 & 79.3 & 90.3 & 88.2\\
        CASIA2 & IRRG & \textbf{93.2} & \textbf{96.0} & \textbf{84.7} & 89.9 & 86.7 & 91.1 & 90.1\\
%         BKHN\_10 & IRRG+DSM+nDSM & 92.9 & 96.0 & 84.6 & 89.8 & 88.8 & 91.0 & 90.4\\
%         \hline
%         FCN \citep{sherrah2016fully} & IRRG & 90.5 & 93.7 & 83.4 & 89.2 & 72.6 & 89.1 & 85.6\\
%         FCN \citep{marmanis2018classification} & IRRG+DSM & 92.3 & 95.2 & 84.1 & 90.0 & 79.3 & 90.3 & 88.2\\
%         SegNet \citep{audebert2018beyond} & IRRG & 91.5 & 94.3 & 82.7 & 89.3 & 85.7 & 89.4 & 88.7\\
%         V-FuseNet \citep{audebert2018beyond} & IRRG+DSM & 91.0 & 94.4 & 84.5 & 89.9 & 86.3 & 90.0 & 89.2\\
        TreeUNet & IRRG+DSM& 92.5 & 94.9 & 83.6 & 89.6 & 85.9 & 90.4 & 89.3 \\
        \hline
%         //92.1 & 96.3 & 79.1 & 90.2 & 83.3
%         Ours (baseline) & IRRG & *** & 89.1 & 86.8\\
        Ours & IRRG+DSM(s) & 93.1 & 95.9 & 84.6 & \textbf{90.1} & \textbf{88.4} & \textbf{91.3} & \textbf{90.4}\\
        \hline
    \end{tabular}
    
    \label{tab_result_vaihingen}
\end{table*}

\subsection{Results on Potsdam}

\begin{figure*}[!t]
\centering
\includegraphics[width=17cm]{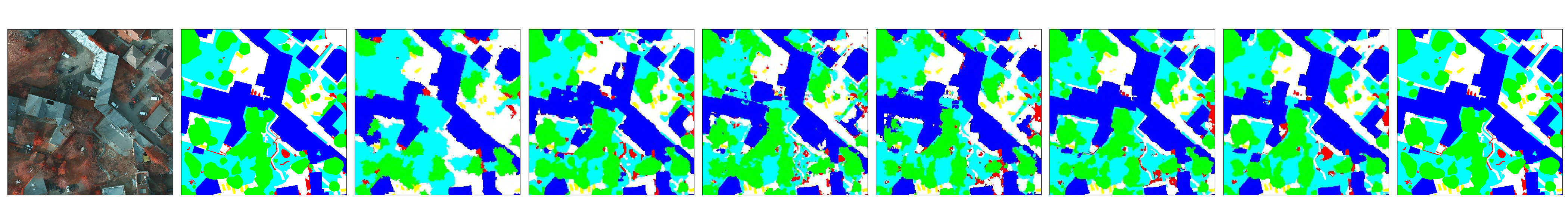}\hspace{2pt}
\includegraphics[width=17cm]{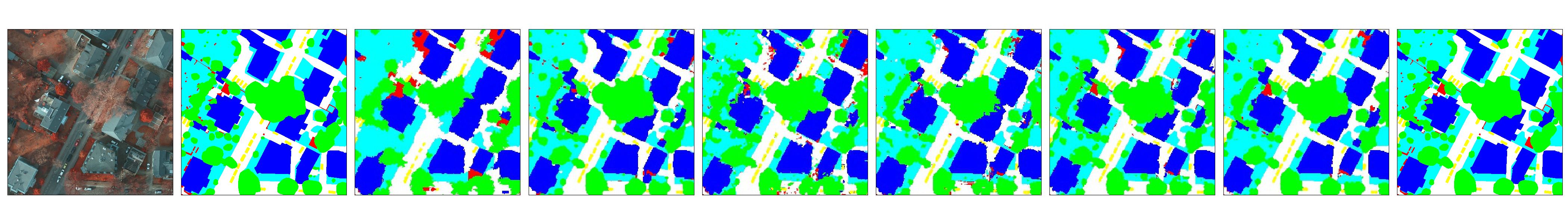}\hspace{2pt}
\includegraphics[width=17cm]{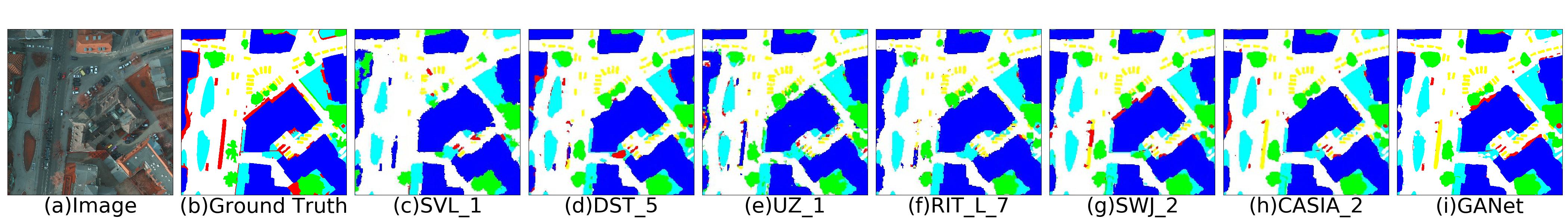}\hspace{2pt}

\caption{Selected examples of the segmentation results on ISPRS Potsdam dataset.}
\label{fig_rst_potsdam}
\end{figure*}

We report the performance of our GANet method and the comparing methods on the ISPRS Potsdam test set in Table \ref{tab_result_potsdam}. As shown in Table \ref{tab_result_potsdam}, our GANet method obtains the best performance on average F1 score and the second-best performance on OA. We note that SWJ\_2 model obtains better performance than our GANet model on OA, and CASIA\_2 obtains quite close performance compared to our model. This is probably because these two comparing models are pre-trained on PASCAL VOC 2012 dataset and then finetuned on the ISPRS Potsdam dataset. In contrast, our GANet model does not require additional datasets for pretraining. One should also note that OA is sensitive to the class distribution, while F1-score is a better metric when there are imbalanced classes as in the above case. Moreover, our model gets new-state-of-the-art performance on 4 out of 5 categories, including building, low vegetation, tree and car.

% Besides, our GANet model only uses simple data augmentation strategy (i.e., randomly horizontal and vertical flip), while  SWJ\_2 and CASIA\_2 models use more complicated data augmentation techniques. 
%From Table \ref{tab_result_potsdam}, one can also find out that those methods using DSM information outperform their counterparts with only optical images as inputs. This demonstrates that using geometric information from DSM images can generally improve the performance of semantic labeling.

Moreover, unlike previous methods (e.g., such as One\_7 \citep{audebert2018beyond} and TreeUNet \citep{yue2019treeunet}) that mostly use elevation data (DSM and nDSM) as additional inputs, our GANet instead uses DSMs as side supervision during training. In the inference stage, our model only needs optical images as inputs and can surprisingly get better performance than previous methods.

We show some examples of the semantic labeling results of our GANet model and the compared methods on the ISPRS Potsdam dataset in Figure \ref{fig_rst_vaihingen}. As shown in this figure, our GANet model obtains reasonable results on the ISPRS Potsdam dataset, which again demonstrates the effectiveness of our GANet model for aerial image classification. A detailed investigation of the qualitative results also shows the advantage of our GANet model for distinguishing objects with similar 2D appearance but different 3D geometric proprieties.

\begin{table*}[!h]
    \centering
    \caption{Classification performance on the Potsdam dataset. `DSM(s)' indicates the model using DSM as additional supervision.}
    \begin{tabular}{c c c c c c c c c}
        \hline
        Method & Input &Imp. surf. & Buildings &    Low veg. &    Trees &    Cars &    OA & Average F1 \\
        \hline
        SVL\_1 &  IRRG+nDSM+NDVI & 83.5 & 91.7 & 72.2 & 63.2 & 62.2 & 77.8 & 74.6 \\
        DST\_5 & IRRGB+DSM & 92.5 & 96.4 & 86.7 & 88.0 & 94.7 & 90.3 & 91.7 \\
        UZ\_1 & IRRG+nDSM & 89.3 & 95.4 & 81.8 & 80.5 & 86.5 & 85.8 & 86.7 \\
        RIT\_L7 & IRRGB+nDSM+NDVI & 91.2 & 94.6 & 85.1 & 85.1 & 92.8 & 88.4 & 89.8 \\
        SWJ\_2 & IRRG & \textbf{94.4} & 97.4 & 87.8 & 87.6 & 94.7 & \textbf{91.7} & 92.4 \\
        CASIA\_2 & IRRGB & 93.3 & 97.0 & 87.7 & 88.4 & 96.2 & 91.1 & 92.5 \\
%         FCN \citep{liu2017dense} & IRRG & 91.2 & 94.6 & 85.1 & 85.1 & 92.8 & 88.4 & 89.8\\
%         FCN \citep{sherrah2016fully} & IRRG & 92.5 & 96.4 & 86.7 & 88.0 & 94.7 & 90.3 & 91.7\\
%         SegNet \citep{audebert2018beyond} & IRRGB & 92.4 & 95.8 & 86.7 & 87.4 & 95.1 & 90.0 & 91.5\\
%         V-FuseNet \citep{audebert2018beyond} & IRRGB+DSM+nDSM & 92.7 & 96.3 & 87.3 & 88.5 & 95.4 & 90.6 & 92.0\\
        TreeUNet & IRRGB+DSM+nDSM & 93.1 & \textbf{97.3} & 86.8 & 87.1 & 95.8 & 90.7 & 92.0 \\
        \hline
        Ours & IRRG+DSM(s) & 93.0 & \textbf{97.3} & \textbf{88.2} & \textbf{89.5} & \textbf{96.8} & 91.3 & \textbf{93.0}\\
%         Ours (baseline) & IRRGB & \\
        % Ours & IRRGB+DSM(s) \\
        \hline
    \end{tabular}
    
    \label{tab_result_potsdam}
\end{table*}

\section{Discussion}\label{sc_discussion}
In this section, we conduct experiments to validate the effectiveness of our proposed height supervision module and the geometry-aware convolution module. We also explore the performance of our GANet model with different network depth. All performances are reported on the validation set of the ISPRS Vaihingen dataset.

\subsection{Effect of Height Supervision}\label{sc_diss_hs}
First, we investigate the effectiveness of height supervision in comparison to methods that directly using DSM images as model input. To achieve this, we remove the height decoder branch as well as the GAC module from our GANet model and use it as a baseline model. The baseline model now becomes a traditional single-stream encoder-decoder network. We report the performance of the baseline and our GANet model with the height decoder branch in Table \ref{tab_diss1}. Moreover, we also list the performance of our GANet model with different configurations of $\lambda$. Note that all models do not include the GAC module in this section.

From Table \ref{tab_diss1}, one can find out that by using the proposed height decoder branch, our GANet model gets a significant improvement, which demonstrates the benefits of using height information as side supervision. Specifically, the baseline model obtains an OA of 90.7\% and an average F1 score of 88.2\% on the Vaihingen validation set, while our GANet model with the height decoder branch achieves an OA of 91.3\% and an average F1 score of 89.6\% when $\lambda$ equals 1. Moreover, Table \ref{tab_diss1} also show that different values of $\lambda$ give similar performance. Our GANet model gets the best performance when $\lambda$ equals 1.

\begin{figure*}[!h]
\centering
\subfigure[Aerial image.]{\includegraphics[width=4.3cm]{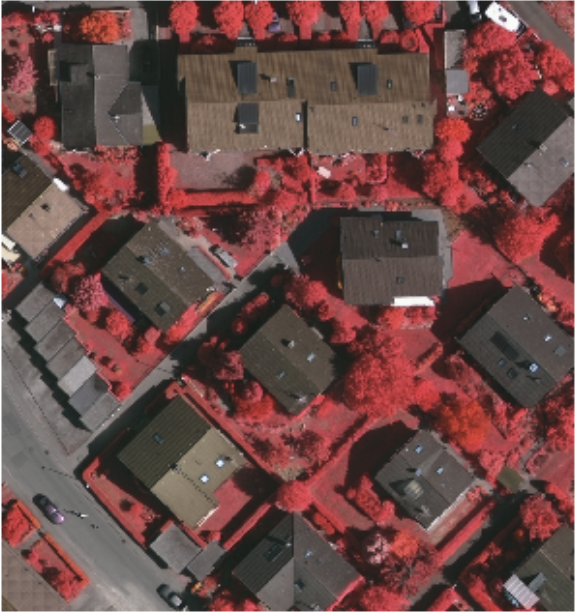}}\hspace{2pt}
\subfigure[DSM.]{\includegraphics[width=4.3cm]{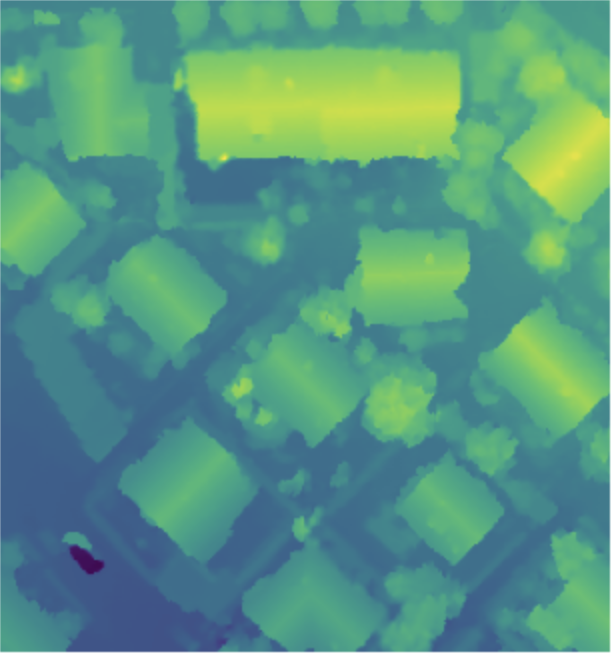}}\hspace{2pt}
\subfigure[Ground Truth.]{\includegraphics[width=4.3cm]{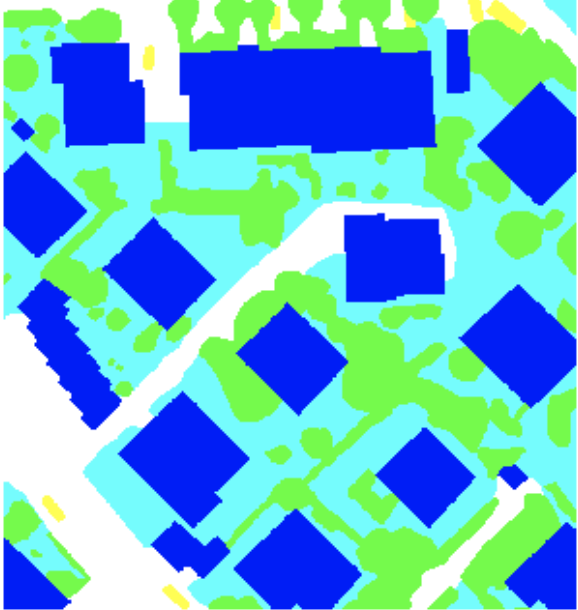}} \\
\vspace{2pt}
\subfigure[GANet* w/o GAC.]{\includegraphics[width=6cm]{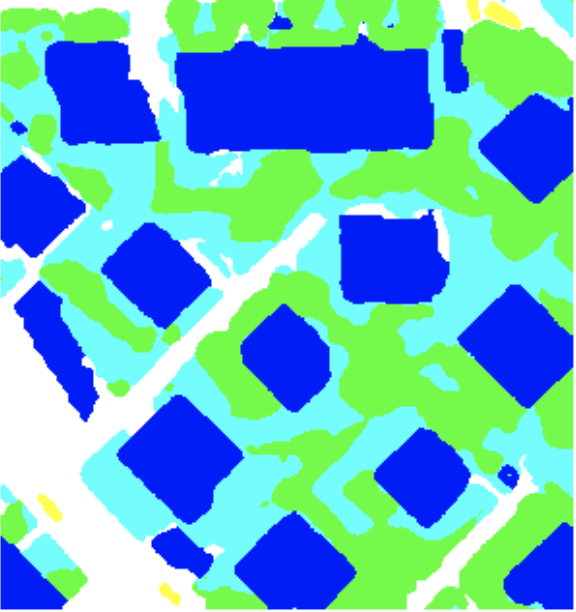}}\hspace{2pt}
\subfigure[GANet ($\lambda=1$) w/o GAC.]{\includegraphics[width=6cm]{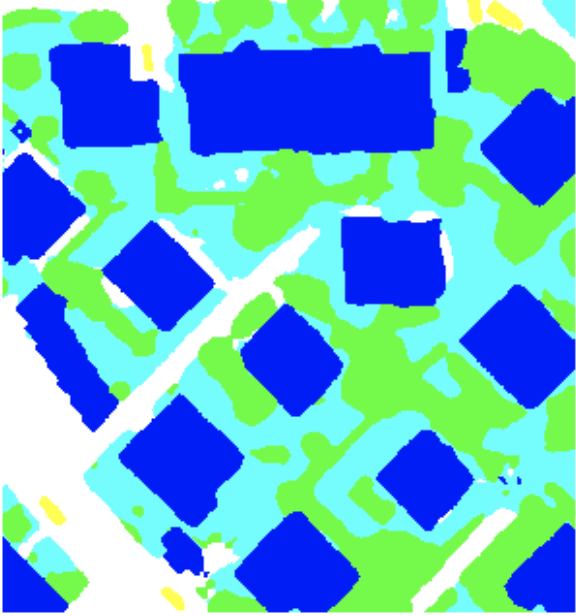}}

\caption{Effect of height supervision on an example of the ISPRS Vaihingen dataset.}
\label{fig_diss1}
\end{figure*}

Figure \ref{fig_diss1} gives an example of the classification results with and without the height estimation branch. As shown in this figure, by incorporation geometrical information from DSM images, our model successfully identifies building pixels that have a similar 2D appearance with impervious surface pixels. Moreover, benefiting from geometrical information, our model also achieves higher accuracy on vegetation-tree classification by distinguishing them using the height information.

\begin{table}[!h]
    \centering
    \caption{Effect of height supervision. Performances are reported on the Vaihingen validation set. GANet* denotes our baseline model without height estimation.}
    \begin{tabular}{c c c c c c c}
        \hline
        Method & OA & Average F1 \\
        \hline
        GANet* w/o GAC & 90.7 & 88.2\\
%         GANet ($\lambda=0.5$) w/o GAC &  91.5 & 89.8\\
%         GANet ($\lambda=1$) w/o GAC & \textbf{91.6} & \textbf{90.1} \\
%         GANet ($\lambda=2$) w/o GAC & 91.5 & 89.6 \\
        GANet ($\lambda=0.5$) w/o GAC &  91.2 & 89.3\\
        GANet ($\lambda=1$) w/o GAC & \textbf{91.3} & \textbf{89.6} \\
        GANet ($\lambda=2$) w/o GAC & 91.2 & 89.1 \\
        \hline
    \end{tabular}
    
    \label{tab_diss1}
\end{table}

\subsection{Effect of GAC Module}\label{sc_diss_gac}
Then, we explore the benefits of using our newly proposed GAC module for contextual and geometrical feature fusion. We investigate the performance of our model with and without the GAC module with $\lambda$ set to 1. We also explore another variant of our GANet model using element-wise summation for feature fusion instead of the proposed GAC module. Table \ref{tab_diss_gac} lists the quantitative performance of our GANet model as well as the comparing methods. The results show that both the element-wise summation fusion strategy and the GAC module can improve the segmentation performance. More importantly, the proposed GAC module performs better than an element-wise summation fusion strategy. This is because our GAC module can effectively learn geometric affinity from the geometrical embeddings and use it to weight the convolutional kernels.

\begin{figure*}[!h]
\centering
\subfigure[Aerial image.]{\includegraphics[width=4.3cm]{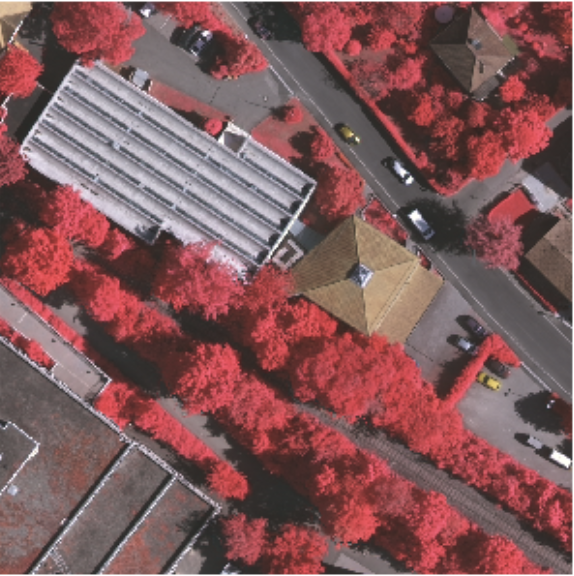}}\hspace{2pt}
\subfigure[DSM.]{\includegraphics[width=4.3cm]{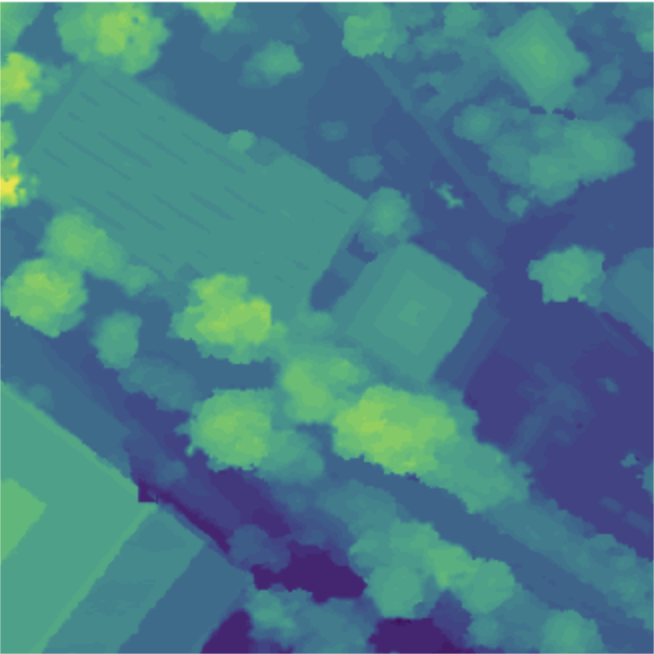}}\hspace{2pt}
\subfigure[Ground Truth.]{\includegraphics[width=4.3cm]{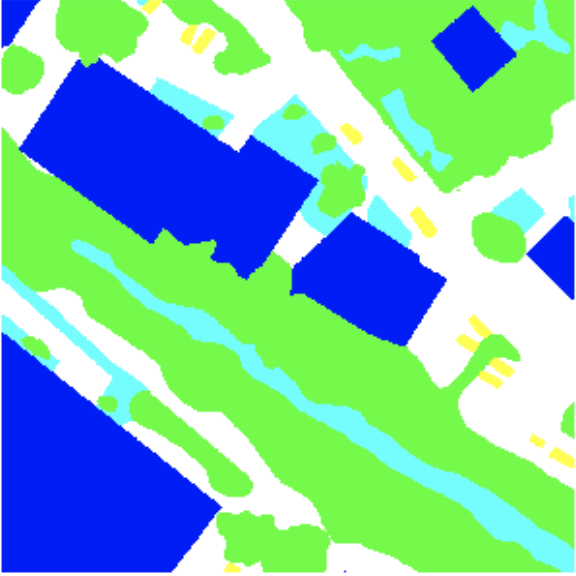}}\hspace{2pt} \\
\vspace{2pt}
\subfigure[GANet w/o feature fusion.]{\includegraphics[width=4.3cm]{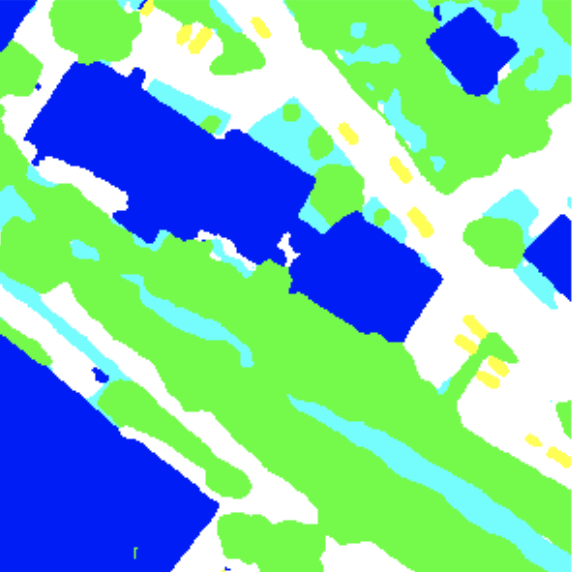}}\hspace{2pt}
\subfigure[GANet w/ Element-wise Summation Fusion.]{\includegraphics[width=4.3cm]{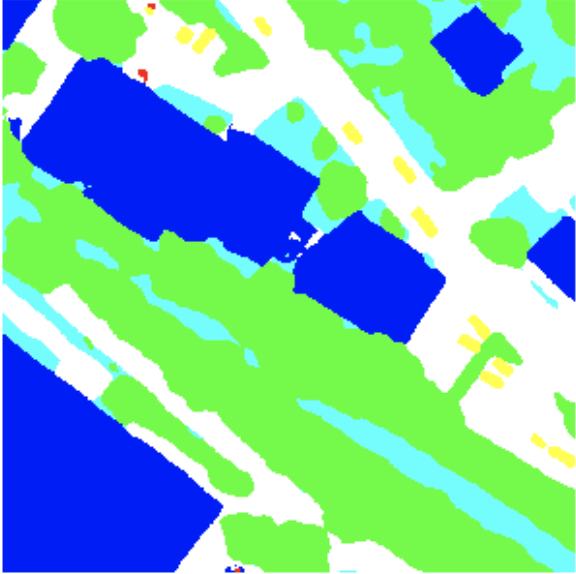}}\hspace{2pt}
\subfigure[GANet w/ GAC fusion]{\includegraphics[width=4.3cm]{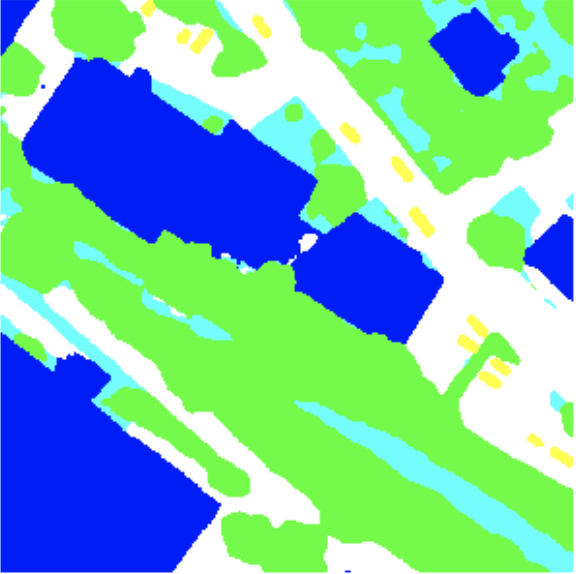}}\hspace{2pt}

\caption{Effect of feature fusion on an example of the ISPRS Vaihingen dataset.}
\label{fig_diss2}
\end{figure*}

Figure \ref{fig_diss2} shows an example of the semantic segmentation results with different fusion strategies. As can be seen in Figure \ref{fig_diss2}, the model without feature fusion misclassifies some building pixels as impervious surface and leads to incorrect classification between low vegetation and tree categories. The two models with feature fusion modules can successfully correct the errors between the building and impervious surface categories using the geometrical information from the height decoder branch. Moreover, our model using the GAC module obtains better performance than its counterpart using an element-wise summation fusion strategy. The improvement mainly comes from a better classification between tree and low vegetation categories.

Furthermore, we compare the performance of our GANet model with another two-stream network call V-FuseNet \citep{audebert2018beyond}. This method uses DSM images as model inputs and fuses the feature embeddings from two separate encoder networks. From Table \ref{tab_diss_gac} one can see that the feature fusion strategy in \citep{audebert2018beyond} obtains a performance boost of  0.9\% on the OA and 1.0\% on the average F1 score. While our GANet model achieves an improvement of 1.3\% on OA and 2.5\% on the average F1 score by using the height supervision and GAC module. This demonstrates that the proposed height supervision and GAC module can make better use of the geometrical information from DSM images. Note that our GANet model does not require DSM images in the test stage, while V-FuseNet \citep{audebert2018beyond} needs DSM images as model input both in the training and test stage. This further demonstrates the superiority of our GANet model over V-FuseNet \citep{audebert2018beyond}.

Moreover, we also investigate the effectiveness of a multi-scale test strategy. As can be seen in Table \ref{tab_diss_gac}, our GANet model enjoys a further performance boost on both evaluation metrics by using a multi-scale test strategy.

\begin{table}[!h]
    \centering
    \caption{Effect of GAC module. Performances are reported on the Vaihingen validation set. 'Sum Fusion' denotes our model using element-wise summation for feature fusion. `ms test' denotes multiscale test.}
    \begin{tabular}{c c c c c c c}
        \hline
        Method & OA & Average F1 \\
        \hline
        SegNet \citep{audebert2018beyond} & 90.2 & 89.3 \\
        V-FuseNet \citep{audebert2018beyond} & 91.1 & 90.3\\
        \hline
        GANet w/o feature fusion & 91.3 & 89.6 \\
        GANet w/ Sum Fusion & 91.6 & 90.1 \\
        GANet w/ GAC & \textbf{92.0} & \textbf{90.7} \\
        \hline
        GANet w/ GAC + ms test  & \textbf{92.3} & \textbf{91.1} \\
        \hline
    \end{tabular}
    
    \label{tab_diss_gac}
\end{table}

\subsection{Effect of Network Depth}\label{sc_diss_depth}
Moreover, we explore the performance of our GANet model with different network depth. Table \ref{tab_net_depth} lists the performance of our GANet model using ResNet-50, ResNet-101 and ResNet-152. Detailed configurations of these architectures can be found in \citep{he2016deep}. As illustrated in Table \ref{tab_net_depth}, our GANet model obtains the best performance when using ResNet-101 architecture. Our mode with ResNet-50 architecture leads to inferior performance due to limited feature extraction abilities, while our model with ResNet-152 architecture suffers from the over-fitting problem, and the performance decreases.

\begin{table}[!h]
    \centering
    \caption{Semantic segmentation performance on ISPRS Vaihingen validation set with different network depths.}
    \begin{tabular}{c c c c c }
        % \hline
        % dataset & \multicolumn{2}{c}{Vaihingen} & \multicolumn{2}{c}{Potsdam} \\
        \hline
        Method & OA & Average F1 \\
        \hline
        ResNet-50 & 91.5 & 89.8\\
        ResNet-101 & \textbf{92.3} & \textbf{91.1} \\
        ResNet-152 & 92.0 & 90.6 \\
        \hline
    \end{tabular}
    
    \label{tab_net_depth}
\end{table}

\subsection{Height Estimation Performance}
\begin{figure*}[t]
\centering
\includegraphics[width=4.3cm]{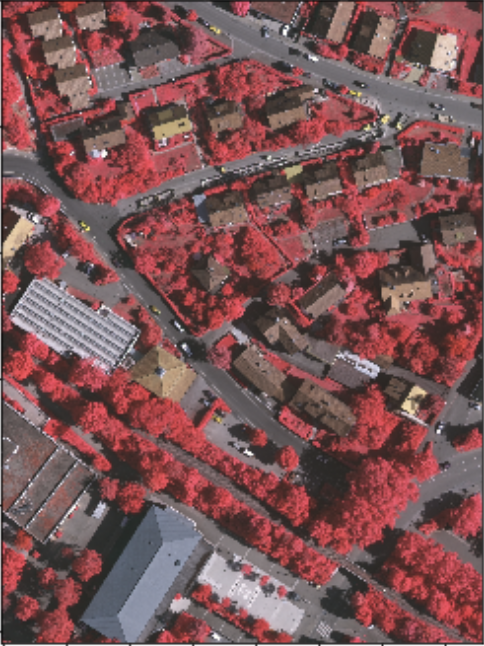}\hspace{2pt}
\includegraphics[width=4.3cm]{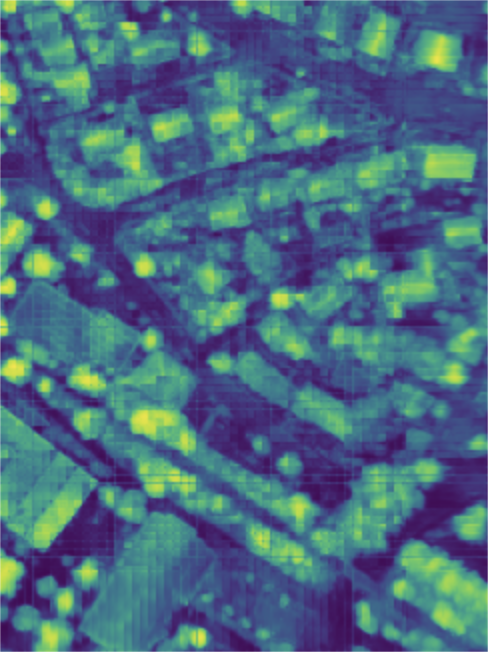}\hspace{2pt}
\includegraphics[width=4.3cm]{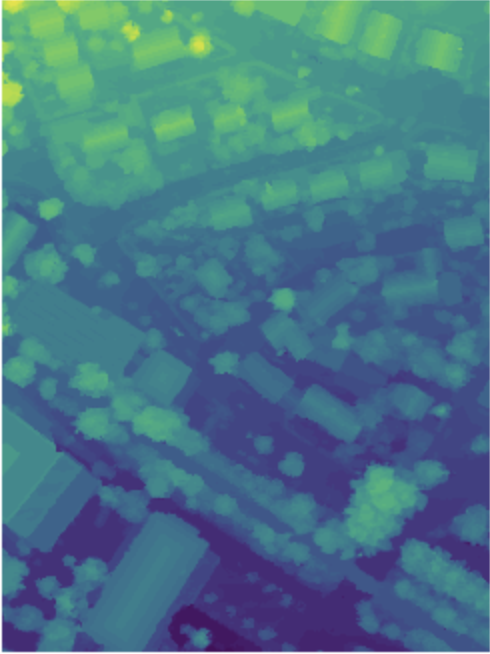}\hspace{2pt}
% \vspace{3pt}
% \subfigure[Aerial images.]{\includegraphics[width=4.3cm]{fig_ab3_img2.png}}\hspace{2pt}
% \subfigure[Predicted height maps.]{\includegraphics[width=4.3cm]{fig_ab3_pred2.png}}\hspace{2pt}
% \subfigure[Ground truth DSMs.]{\includegraphics[width=4.3cm]{fig_ab3_gt2.png}}\hspace{2pt}

\caption{Height estimation on ISPRS Vaihingen dataset.}
\label{fig_diss3}
\end{figure*}

To demonstrate that our GANet model can learn geometrical features and predict height maps, we visualized the predicted height maps on the ISPRS Vaihingen validation set in Figure \ref{fig_diss3}. As shown in Figure \ref{fig_diss3}, our GANet model obtains satisfying height estimation performance on the ISPRS Vaihingen dataset. Specifically, our GANet model predicts larger height values for building and tree pixels while it predicts smaller height values for those pixels from the impervious surface, low vegetation, and car categories. Note that our GANet model only predicts the relative height values in the range of [0,1], while the ground truth height maps show real-value altitude. For example, there is a height lifting from bottom to top in the ground truth DSM image in Figure \ref{fig_diss3}. One can use post-processing techniques (e.g., \citep{amirkolaee2019height}) to merge the predicted height patches and produce absolute height maps.

\section{Conclusions}\label{sc_conclusion}
In this paper, we introduce a geometry-aware convolutional neural network to approach the problem of semantic segmentation of remote sensing images. Our model benefits from the 3D geometric information via joint height estimation. Unlike previous methods that mostly use a single decoder network to predict pixel-wise semantic labels, in our model, a newly designed height decoder branch is developed to predict the height map under the supervision of DSM images. The height decoder branch is trained to be capable of distilling 3D geometric features from 2D contextual/appearance features. In this way, our model does not require DSM as model input and still benefits from the helpful 3D geometric information. Furthermore, we introduce a novel geometry-aware convolution module to combine the learned 3D geometric features and 2D contextual features from two decoder branches. With the fused feature embeddings, our model can produce geometry-aware segmentation with enhanced performance. In the training stage, our model uses DSM images as side supervision to enforce geometric feature distillation, while in the inference stage, it does not need DSM data and can directly produce the segmentation labels. Experiments on ISPRS Vaihingen and Potsdam datasets demonstrate the effectiveness of our proposed method for aerial image classification. Our proposed method achieves remarkable performance on these two datasets without using any hand-crafted features or post-processing.

\section*{ACKNOWLEDGEMENTS}\label{ACKNOWLEDGEMENTS}
% We would like to gratefully acknowledge the ISPRS for providing the experimental dataset.
We would like to acknowledge the German Society for Photogrammetry, Remote Sensing and Geoinformation (DGPF) (\url{http://www.ifp.uni-stuttgart.de/dgpf/DKEP-Allg.html}) for providing the Vaihingen dataset. The authors thank the ISPRS WG II/4 for releasing the Vaihingen and Potsdam datasets and organizing the 2D semantic labeling contest.

% biography section
% 
% If you have an EPS/PDF photo (graphicx package needed) extra braces are
% needed around the contents of the optional argument to biography to prevent
% the LaTeX parser from getting confused when it sees the complicated
% \includegraphics command within an optional argument. (You could create
% your own custom macro containing the \includegraphics command to make things
% simpler here.)
%\begin{IEEEbiography}[{\includegraphics[width=1in,height=1.25in,clip,keepaspectratio]{mshell}}]{Michael Shell}
% or if you just want to reserve a space for a photo:
{%\footnotesize

    \begin{spacing}{0.9}% tune the size by altering the parameter
        \bibliography{bibliography} % Include your own bibliography (*.bib), style is given in isprs.cls
    \end{spacing}
}

\end{document}